\documentclass[journal]{IEEEtran}
\ifCLASSINFOpdf
   \usepackage[pdftex]{graphicx}
\else
   \usepackage[dvips]{graphicx}
\fi
\usepackage{amsmath}
\usepackage{amsfonts}
\usepackage{amssymb}
\usepackage{algorithmic}
\usepackage{algorithm}

\makeatletter
\newcommand{\removelatexerror}{\let\@latex@error\@gobble}
\makeatother

\usepackage{booktabs}
\usepackage{multirow}
\newcommand{\ie}[1]{\emph{i.e.}#1}

\begin{document}
%
\title{LIF-Seg: LiDAR and Camera Image Fusion for 3D LiDAR Semantic Segmentation}
%
%
%


\author{Lin~Zhao, Hui~Zhou, Xinge~Zhu, Xiao~Song, Hongsheng~Li, Wenbing~Tao 
	\thanks{L. Zhao and W. Tao are with the National Key Laboratory of Science and Technology on Multispectral Information Processing, School of Artifical Intelligence and Automation, Huazhong University of Science and Technology, Wuhan 430074, China. E-mail: linzhao@hust.edu.cn, wenbingtao@hust.edu.cn. (Corresponding author: Wenbing~Tao)}
	\thanks{H. Zhou and X. Song are with SenseTime Research. E-mail: smarthuizhou@gmail.com, songxiao@sensetime.com.}
	\thanks{X. Zhu and H. Li are with the Chinese University of Hong Kong. E-mail: zhuxinge123@gmail.com, lihongsheng@gmail.com.}
	\thanks{Manuscript received XXXX, 2021; revised XXXX, 2021.}
}

%
%

\markboth{IEEE Transactions on Multimedia,~Vol.~XX, No.~XX, XXXX~2021}%
{}
%



\maketitle

\begin{abstract}
Camera and 3D LiDAR sensors have become indispensable devices in modern autonomous driving vehicles, where the camera provides the fine-grained texture, color information in 2D space and LiDAR captures more precise and farther-away distance measurements of the surrounding environments. The complementary information from these two sensors makes the two-modality fusion be a desired option. However, two major issues of the fusion between camera and LiDAR hinder its performance, \ie, how to effectively fuse these two modalities and how to precisely align them (suffering from the weak spatiotemporal synchronization problem). In this paper, we propose a coarse-to-fine LiDAR and camera fusion-based network (termed as LIF-Seg) for LiDAR segmentation. For the first issue, unlike these previous works fusing the point cloud and image information in a one-to-one manner, the proposed method fully utilizes the contextual information of images and introduces a simple but effective early-fusion strategy. Second, due to the weak spatiotemporal synchronization problem, an offset rectification approach is designed to align these two-modality features. The cooperation of these two components leads to the success of the effective camera-LiDAR fusion. Experimental results on the nuScenes dataset show the superiority of the proposed LIF-Seg over existing methods with a large margin. Ablation studies and analyses demonstrate that our proposed LIF-Seg can effectively tackle the weak spatiotemporal synchronization problem. 
\end{abstract}


%
\IEEEpeerreviewmaketitle

\section{Introduction}
\IEEEPARstart{W}{ith} the rapid development of autonomous driving, 3D scene perception has received more and more attention in recent years, especially in computer vision and deep learning. LiDAR has become an indispensable 3D sensor in autonomous driving. Point clouds acquired by LiDAR, compared with data from other sensors (e.g., cameras and radars), can provide rich geometric, scale information, accurate distance measurements and fine semantic descriptions, which are quite helpful in understanding 3D scenes for autonomous driving planning and execution.

\begin{figure}[!t]
	\centering
	\includegraphics[width=0.95\columnwidth]{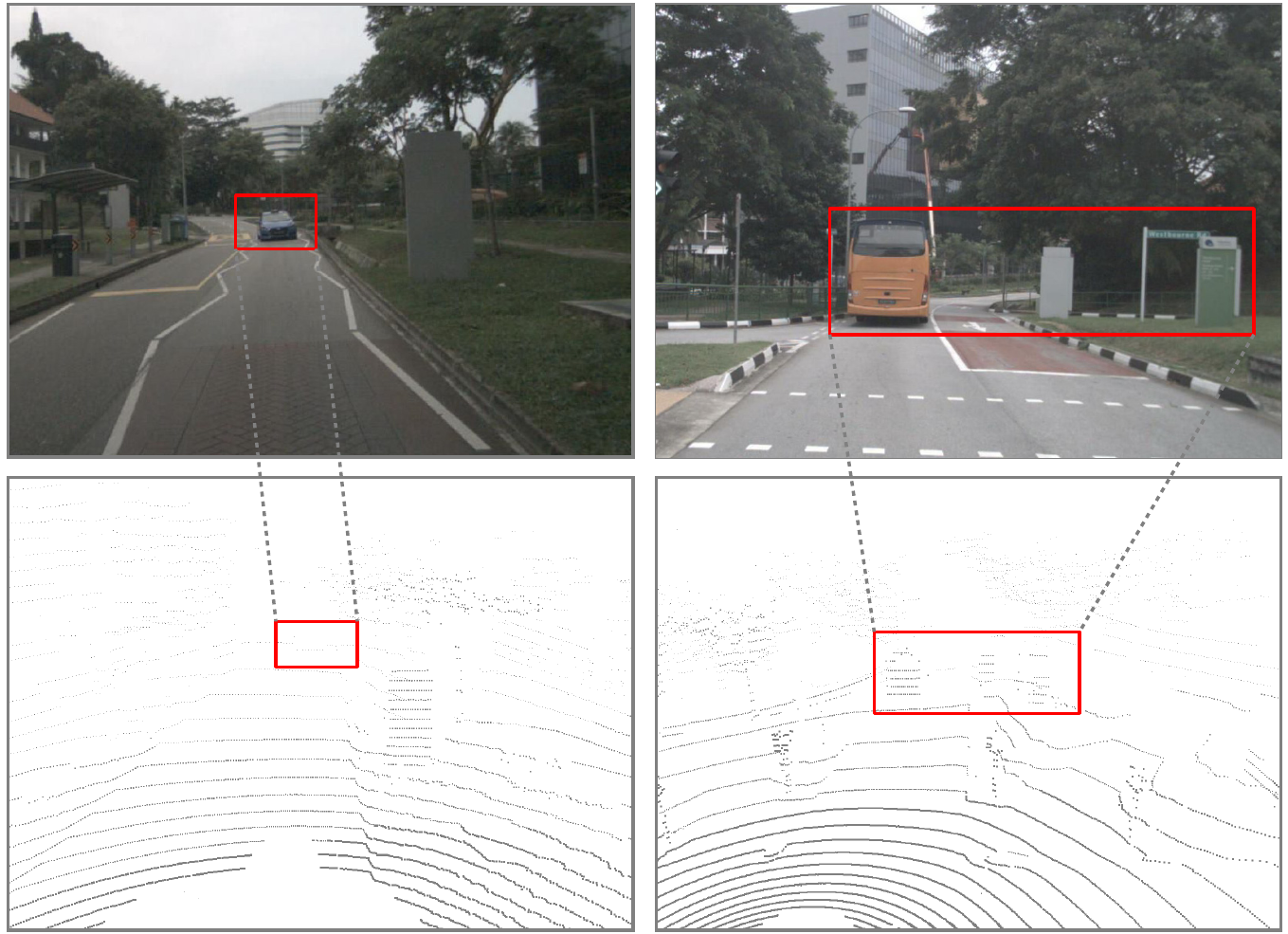}
	\caption{Example scenes from the nuScenes \cite{caesar2020nuscenes} dataset. In the left case, the LiDAR points become too sparse and difficult to identify a car in the distance. The right case shows that the bus and two constructions appear very similar in the point cloud, which makes it difficult for the segmentation model to distinguish the bus from the background. }
	\label{fig:lcf_description}
\end{figure}

LiDAR point cloud semantic segmentation aims to assign a special semantic category for each 3D point, which is a critical task for autonomous driving. This task can help the perception system to recognize and locate dynamic objects and drivable surfaces. Although the classic task of 3D object detection has developed relatively mature solutions \cite{lang2019pointpillars,shi2020pv,zhang2020spatial} to support real-world autonomous driving, it has difficulty to recognize and locate the drivable surfaces. In general, LiDAR point clouds are sparse, and their sparseness usually increases as the reflection distance increases, which makes it difficult for the semantic segmentation model to segment small objects in the distance, as illustrated in the left of Fig. \ref{fig:lcf_description}.

As mentioned above, although LiDAR points can provide accurate distance measurements and capture the structures of objects, they are usually sparse, unordered, and unevenly distributed. Recently, some methods \cite{kochanov2020kprnet,zhou2020cylinder3d,cheng2021af,xu2021rpvnet} based only on LiDAR have significantly improved the performance of 3D semantic segmentation, but the performance of these methods are still limited because of lacking dense and rich information on the objects such as their colors, and textures, as depicted in the right of Fig. \ref{fig:lcf_description}. Compared with point clouds, camera images contain more regular and dense pixels and have much richer semantic information (e.g., color, texture) to distinguish different semantic categories, while suffering from the lack of depth and scale information. Therefore, the complementary information from LiDAR and camera makes the two modalities fusion be a desired option. However, how to effectively fuse these two modalities so that we can make full use of the advantages of these two sensors to generate better and more reliable accurate semantic segmentation results.

\begin{figure}[!t]
	\centering
	\includegraphics[width=0.98\columnwidth]{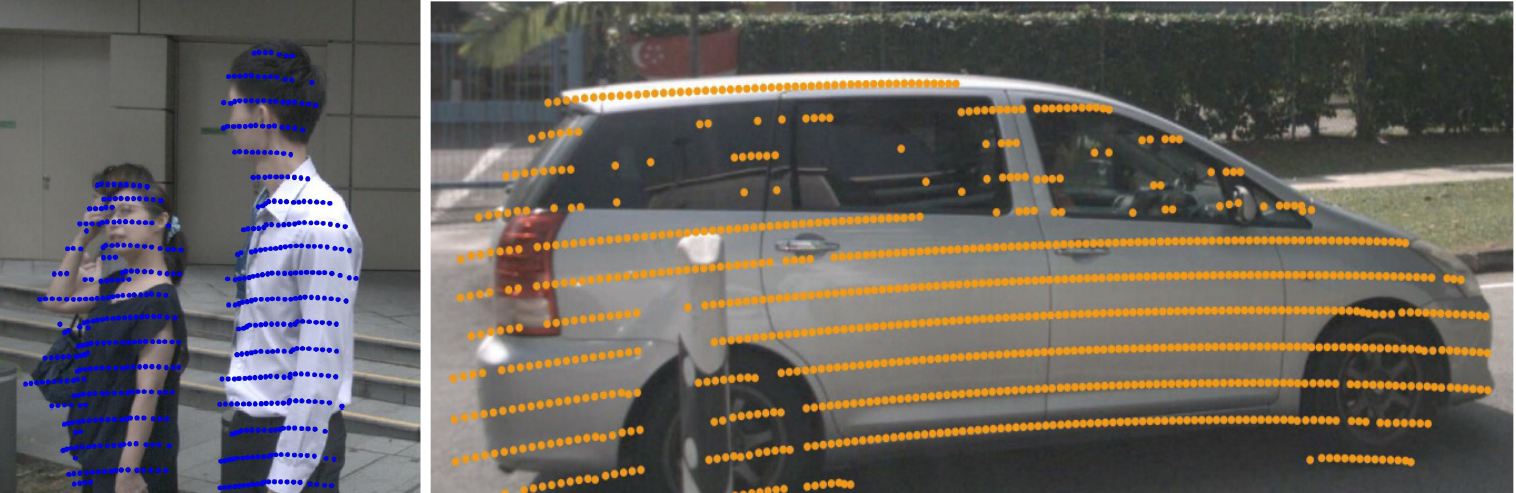}
	\caption{Example scenes from the nuScenes \cite{caesar2020nuscenes} dataset. The LiDAR points are projected into the camera image, and some of the projected points fall outside the instance object because of the weak spatiotemporal synchronization problem between the LiDAR and cameras.}
	\label{fig:wss_description}
\end{figure}

Recently, some autonomous driving datasets containing LiDAR point clouds and images have emerged, such as KITTI \cite{geiger2012are} and nuScenes \cite{caesar2020nuscenes}. These datasets not only provide the possibility to combine the advantages of point clouds and images, but also play an important role in promoting the development of point cloud semantic segmentation in academia and industry. However, as illustrated in Fig. \ref{fig:wss_description}, there is a weak spatiotemporal synchronization problem between the LiDAR and the cameras. Some strategies can be used to alleviate this problem. For example, the KITTI and nuScenes realign the point clouds and images with time-stamped sensor metadata, but there is still a certain deviation. The weak spatiotemporal synchronization problem also limits the performance of the fusion between the camera and LiDAR.

Motivated by above findings, we propose a coarse-to-fine framework, named LIF-Seg, to fuse the LiDAR and camera for 3D LiDAR point cloud semantic segmentation. For the first issue, unlike these previous works fusing the point cloud and image information in a one-to-one manner, in the coarse stage, LiDAR points are projected into each camera image, and the $3\times 3$ contextual information of each pixel is concatenated to the intensity measurement of the LiDAR points. The concatenated LiDAR points are fed into a UNet segmentation sub-network (e.g., Cylinder3D \cite{zhou2020cylinder3d}) to obtain coarse LiDAR point features. For the weak spatiotemporal synchronization problem, an offset rectification approach is designed to align the coarse features and image semantic features. Specifically, an image semantic segmentation sub-network (e.g., DeepLabv3+ \cite{chen2018encoder}) is used to extract image semantic features. The coarse features are projected into each image. The projected coarse features are further fused with image semantic features to predict an offset between each projected point and the corresponding image semantic pixel. The predicted offset is used to compensate and align these two-modality features, and then the aligned image semantic features are fused with the coarse features. In the refinement stage, the fused features are fed into a sub-network to refine and generate more accurate predictions. The LIF-Seg not only fuses the LiDAR point features and the different level image features but also effectively tackles the weak spatiotemporal synchronization problem between the LiDAR and cameras.  

The main contributions of this work are as follows: 
(1) We fully utilize the low-level image contextual information and introduce a simple but effective early-fusion strategy. 
(2) We propose an offset rectification method to address the weak spatiotemporal synchronization problem between the LiDAR and cameras. 
(3) We construct a coarse-to-tine LiDAR and camera fusion-based network LIF-Seg for LiDAR semantic segmentation. Experimental results on the nuScenes dataset demonstrate the effectiveness of our method.

\section{Related Work}
In this section, we will briefly review existing works related to our method: deep learning for 3D point clouds, LiDAR point cloud semantic segmentation, LiDAR and camera fusion methods, image semantic segmentation. Especially, we mainly focus on the LiDAR-only and fusion-based methods.

\subsection{Deep learning for 3D Point Clouds}
Different from 2D image processing methods, point clouds processing is a challenging task because of its irregular and unordered properties. PointNet \cite{qi2017pointnet} is one of the first works of directly learning the point features based on the raw point clouds through a shared Multi-Layer Perceptron (MLP) and max-pooling. Some subsequent works \cite{qi2017pointnet++,wu2019pointconv,zhao2020jsnet,engelmann20203d,liu2020semantic,chen2020hapgn,zhang2020pointhop,zhang2020fusion,qiu2021geometric} are often based on the pioneering works (e.g., PointNet, PointNet++) and further promote the effectiveness of sampling, grouping and ordering to improve the performance of semantic segmentation. Other methods \cite{wang2019graph,wang2019dynamic,fu2021dynamic} extract the hierarchical point features by introducing a graph network. Although these methods have achieved promising segmentation results on indoor point clouds, most of them cannot be directly trained or scaled up to large-scale outdoor LiDAR point clouds due to the varying density and large range of scenes. Moreover, a large number of points also cause these methods to have expensive computational and memory consumption when adapting to outdoor scenes.

\subsection{LiDAR Point Cloud Semantic Segmentation}
As the availability of public datasets \cite{caesar2020nuscenes,behley2019semantickitti} increasing, LiDAR point cloud semantic segmentation research is developing. Currently, these methods can be grouped into three main categories: projection-based, voxel-based and multi-view fusion-based methods. 

Projection-based methods focus on mapping the 3D point clouds to a regular and dense 2D image so that 2D CNN can be used to process the pseudo image. SqueezeSeg \cite{wu2018squeezeseg}, SqueezeSegv2 \cite{wu2019squeezesegv2}, RangeNet++ \cite{milioto2019rangenet++}, SalsaNext \cite{cortinhal2020salsanext} and KPRNet \cite{kochanov2020kprnet} utilize the spherical projection mechanism to convert the point clouds into a range image, and adopt an encoder-decoder network to obtain semantic information. For instance, KPRNet \cite{kochanov2020kprnet} presents an improved architecture and achieves promising results by using a strong ResNeXt-101 backbone with an Atrous Spatial Pyramid Pooling (ASPP) block, and it also applies KPConv \cite{thomas2019kpconv} as segmentation head to replace the inefficient KNN postprocessing. PolarNet \cite{zhang2020polarnet} utilizes a polar Birds-Eye-View (BEV) instead of the standard 2D grid-based BEV projections. However, these projection-based methods inevitably loss and alter the original topology, leading to the failure of geometric information modeling.

Voxel-based methods convert point clouds into voxels and then apply vanilla 3D convolutions to obtain segmentation results. More recently, some works \cite{graham20183d,tang2020searching} are proposed to accelerate the 3D convolution, and improve the performance with less computational and memory consumption. Following the previous works \cite{graham20183d,tang2020searching}, 3D-MPA \cite{engelmann20203d}, PointGroup \cite{jiang2020pointgroup} and OccuSeg \cite{han2020occuseg} achieve significant segmentation results on indoor point clouds. As mentioned above, these methods cannot be directly used for outdoor LiDAR point cloud segmentation because of the inherent properties of outdoor point clouds, including sparsity and varying density. Furthermore, Cylinder3D \cite{zhou2020cylinder3d} utilizes cylindrical partition and designs an asymmetrical residual block to further reduce computation.

Multi-view fusion-based methods combine voxel-based, projection-based and/or point-wise operations for LiDAR point clouds segmentation. To extract more semantic information, some recent methods \cite{liu2019point,gerdzhev2020tornado,liong2020amvnet,wang2020pillar,zhou2020end,zhang2020deep,chen2020mvlidarnet,cheng2021af,xu2021rpvnet} blend two or more different views together. For instance, \cite{wang2020pillar,zhou2020end} combine point-wise information from BEV and range-image in early-stage, and then feed it to the subsequent network. AMVNet \cite{liong2020amvnet} utilizes the uncertainty of different view outputs to do late-fusion. PVCNN \cite{liu2019point}, FusionNet \cite{zhang2020deep} and $\left(AF\right)^2$-S3Net \cite{cheng2021af} use point-voxel fusion scheme to achieve better segmentation results. RPVNet \cite{xu2021rpvnet} proposes a deep fusion network to fuse range-point-voxel three views by a gated fusion mechanism. However, the performance of these methods is also limited due to the LiDAR point clouds lacking rich colors and textures.

\subsection{LiDAR and Camera Fusion Methods}
To make full use of the advantages of the camera and LiDAR sensors, some methods \cite{chen2017multi,ku2018joint,liang2018deep,qi2018frustum,vora2020pointpainting,xie2020pi,pang2020clocs,yoo20203d,huang2020epnet} have been proposed for the camera and LiDAR fusion, especially in 3D object detection task. PI-RCNN \cite{xie2020pi} fuses the camera and LiDAR features by conducting point-wise convolution on 3D points and applying a point-pooling with an aggregation operation. CLOCs \cite{pang2020clocs} operates on the combined output candidates before non-maximum suppression of any 2D and any 3D detector. 3D-CVF \cite{yoo20203d} combines the camera and LiDAR features by using a cross-view spatial feature fusion strategy for better detection performance. EPNet \cite{huang2020epnet} proposes a LiDAR guided Image Fusion module to enhance the LiDAR point features with corresponding image semantic features in multiple scales. PointPainting \cite{vora2020pointpainting} projects lidar points into the output of an image-only semantic segmentation network and appends the class scores to each point, and then feeds it to a LiDAR detector. These methods have achieved promising performance in 3D object detection. However, there are a few previous works which focus on 3D semantic segmentation by combining the advantages of camera and LiDAR, and tackle the weak spatiotemporal synchronization problem of sensors between the camera and LiDAR.

\begin{figure}[!h]
	\centering
	\includegraphics[width=0.98\linewidth]{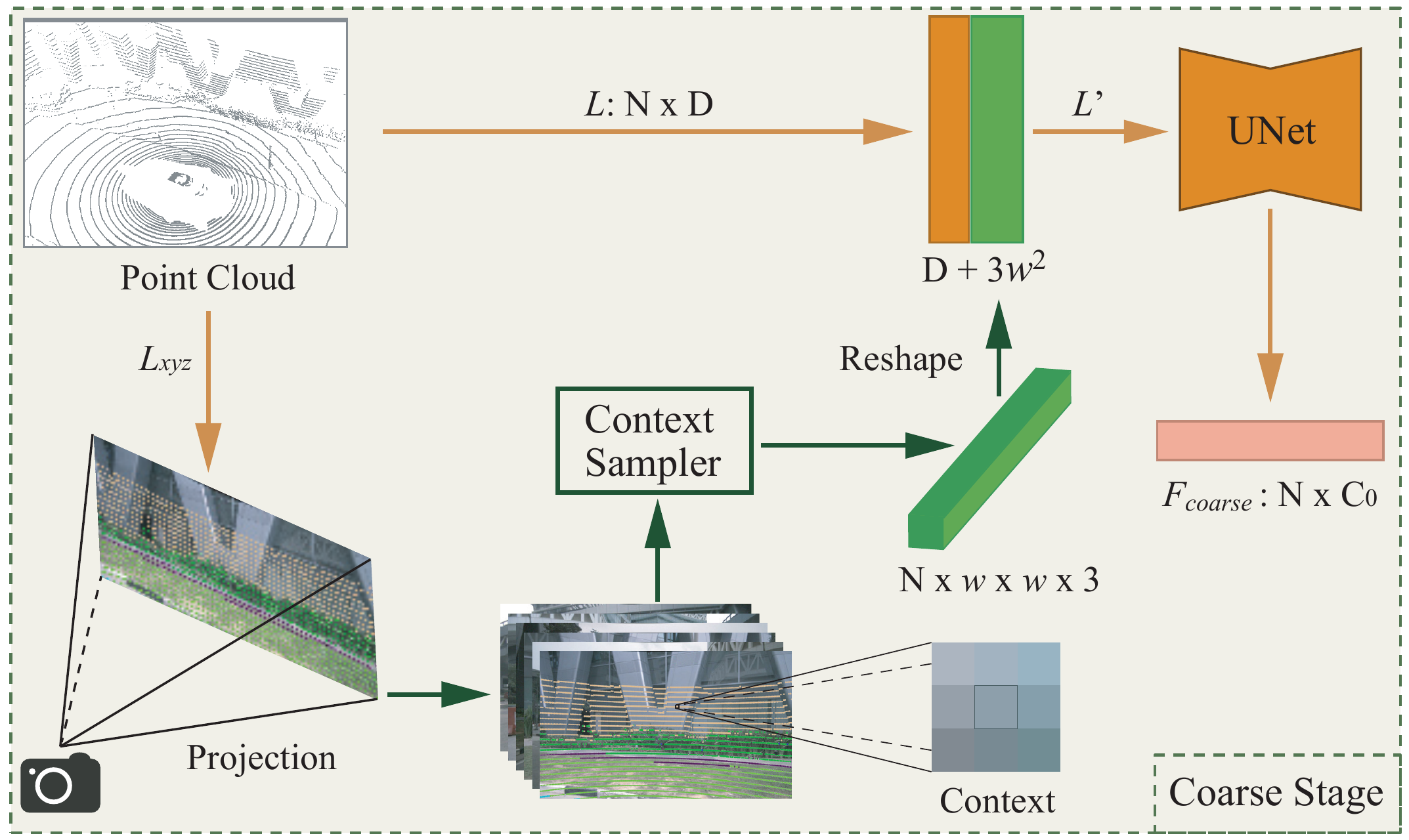}
	\caption{Illustration of coarse feature extraction stage. LiDAR points and low-level image contextual information are fused in this stage. The context sampler is used to extract the image contextual information at the position of each projected point.}
	\label{fig:lifseg_network_context}
\end{figure}

\subsection{Image Semantic Segmentation}
Image semantic segmentation is an important fundamental task in computer vision and has achieved much progress. FCN \cite{long2015fully} is the pioneering work of directly adopting fully convolutional layers to generate image semantic segmentation results. The family of DeepLab \cite{chen2018encoder} utilizes atrous convolution and ASPP modules to capture the contextual information of the image. STDC2 \cite{fan2021rethinking} reduces the inference time-consuming by using a detail guidance module to encode low-level spatial information, but with relatively low performance. Trade-off between efficiency and performance, we adopt DeepLabv3+ \cite{chen2018encoder} as our image segmentation submodel in this work.

\begin{figure}[!t]
	\renewcommand{\algorithmicrequire}{\textbf{Input:}}
	\renewcommand{\algorithmicensure}{\textbf{Output:}}
	\renewcommand{\algorithmiccomment}[1]{\textit{\small\# #1}} 
	\newcommand{\newcomment}[2][3em]{\hskip#1 \algorithmiccomment{#2}} 
	\removelatexerror
	\begin{algorithm}[H]
		\caption{LIF-Seg$\left(L, \mathcal{I}, \mathcal{T}, \mathcal{K}\right)$}
		\label{alg:lifseg}
		\begin{algorithmic}[1]
			\REQUIRE \quad \\
			LiDAR points $L \in \mathbb{R}^{N\times D}$ with $N$ points and $D \geq 3$. \\
			Images $\mathcal{I} = \left\lbrace I_{i} | i=1,2,\dots,n\right\rbrace$ with $n$ cameras. \\
			Transformation matrixes $\mathcal{T}= \left\lbrace T_{i}|i=1,2,\dots,n\right\rbrace$. \\
			Camera matrixes $\mathcal{K}= \left\lbrace K_{i}|i=1,2,\dots,n\right\rbrace$. \\
			Where $I_{i} \in \mathbb{R}^{H\times W\times 3}$, $T_{i} \in \mathbb{R}^{4\times 4}$ and $K_{i} \in \mathbb{R}^{3\times 4}$.
			\ENSURE	\quad \\
			Segmentation scores $S \in \mathbb{R}^{N\times C}$ with $C$ classes.
			\STATE \textit{\small\# Coarse Feature Extraction Stage}
			\STATE Let $Idx = List(),\; Mask = List()$
			\STATE Let $P = Zeros([N, 3w^2])$ with $w\times w$ context.
			\FOR{$i=1$ \TO $n$}
			\STATE $idx = PROJECT\left(K_{i}, T_{i}, L_{xyz}\right)$
			\newcomment[2em]{$idx \in \mathbb{R}^{N\times 2}$}
			\STATE $mask = (0 < idx[:,0] < H \; and \; 0 < idx[:, 1] < W )$
			\STATE $idx = idx[mask, :]$
			\newcomment[4.5em]{$idx \in \mathbb{R}^{N_{i}\times 2}, N_{i} \leq N$}
			\STATE $p = Context(I_{i}, idx, w)$
			\newcomment[4.5em]{$p \in \mathbb{R}^{N_{i}\times w\times w\times 3}$}
			\STATE $P[mask, :] = Reshape(p, [N_{i}, 3w^2])$
			\STATE $Idx.append(idx),\quad Mask.append(mask)$
			\ENDFOR
			\STATE $L^{'} = Concatenate\left(\left[L, P\right], axis=1\right)$
			\STATE $F_{coarse} = UNet_{coarse}(L^{'})$
			\newcomment[4em]{$F_{coarse}\in\mathbb{R}^{N\times C_0}$}
			\STATE \quad
			\STATE \textit{\small\# Offset Learning Stage}
			\STATE $F_{image} = Seg.Net(\mathcal{I})$
			\newcomment[3.5em]{$F_{image}\in\mathbb{R}^{n\times H\times W\times C_1}$}
			\STATE Let $F_{points} = Zeros([n, H, W, C_{0}])$
			\FOR{$i=1$ \TO $n$}
			\STATE $idx = Idx[i]$,\quad $mask = Mask[i]$
			\STATE $F_{points}[i, idx[:, 0], idx[:, 1], :] = F_{coarse}[mask, :]$
			\ENDFOR
			\STATE $F_{\textit{offset}} = Concatenate([F_{image}, F_{points}], axis=3)$
			\STATE $\textit{Offset} = Convs(F_{offset})$
			\newcomment[3.5em]{$\textit{Offset}\in \mathbb{R}^{n\times H\times W\times 2}$}
			\STATE Let $F^{'}_{image} = Zeros([N, C_{1}])$
			\STATE Let $O = Zeros([N, 2])$
			\newcomment[6.4em]{Point-wise offset}
			\FOR{$i=1$ \TO $n$}
			\STATE $idx = Idx[i]$, \quad $mask = Mask[i]$
			\STATE $o = \textit{Offset}[i, idx[:, 0], idx[:, 1], :]$
			\STATE $O[mask, :] = o$
			\STATE $idx = idx + o$ 
			\newcomment[5.4em]{Update the index of points}
			\STATE $F^{'}_{image}[mask,:] = F_{image}[i, idx[:, 0], idx[:, 1], :]$
			\ENDFOR
			\STATE \quad
			\STATE \textit{\small\# Refinement Stage}
			\STATE $F = Concatenate([F_{coarse}, F^{'}_{image}], axis=1)$
			\STATE $S = UNet_{refine}\left(F\right)$ 
			\newcomment[2.2em]{$F\in\mathbb{R}^{N\times (C_{0}+C_{1})}$, $S\in\mathbb{R}^{N\times C}$}
		\end{algorithmic}
	\end{algorithm}
\end{figure}

\section{Proposed Method}
Exploiting the advantages of LiDAR and camera to complement each other is very important for accurate LiDAR point cloud semantic segmentation. However, most existing methods do not make full use of the camera image context information and ignore the weak spatiotemporal synchronization problem between the LiDAR and cameras, limiting the ability of the fusion model to recognize fine-grained patterns. In this paper, we propose a coarse-to-fine framework named LIF-Seg to improve the performance of LiDAR segmentation from two aspects including low-level image contextual information fusion in early-stage and aligned high-level image semantic information fusion in mid-stage. The LIF-Seg accepts LiDAR points and camera images as input and predicts the semantic label of each point. It consists of three main stages: coarse feature extraction stage, offset learning stage and refinement stage. We will give a detailed introduction of those three aspects in the following subsections. %

\subsection{Coarse Feature Extraction Stage}
LiDAR points can provide accurate distance measurements and capture the structures of objects, and camera images contain more regular and dense pixels and have much richer semantic information. Some methods \cite{vora2020pointpainting,xie2020pi,pang2020clocs} attempt to blend LiDAR and camera views together in different stage (e.g., early-fusion, mid-fusion and late-fusion) for 3D object detection. Most of these methods only fuse the low-level or high-level image information in a one-to-one manner. However, the contextual information of the image is also important when fusing the views from LiDAR and the camera. In the coarse stage, we fuse the LiDAR points and low-level image contextual information to obtain the coarse features. 

\begin{figure}[!h]
	\centering
	\includegraphics[width=0.98\linewidth]{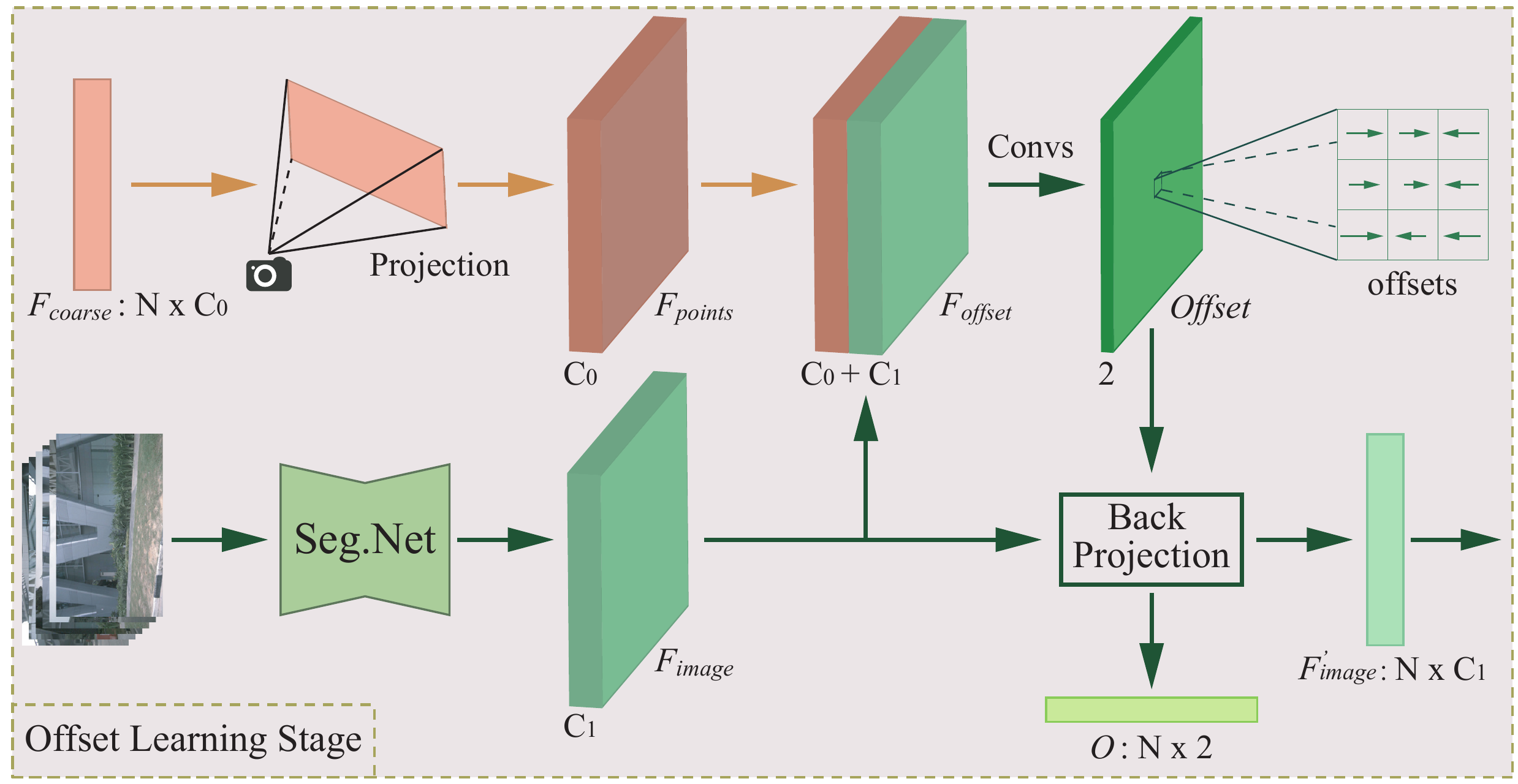}
	\caption{Illustration of offset learning stage. The prediction $\textit{Offset}$ is used to compensate and update the position of projected points in image semantic features. The back-projection operation is used to generate point-wise image semantic features $F^{'}_{image}$ and point-wise offset $O$.}
	\label{fig:lifseg_offset_learning}
\end{figure}

As depicted in Fig. \ref{fig:lifseg_network_context} and Algorithm \ref{alg:lifseg}, each point in LiDAR points $L$ has spatial location $\left(x,y,z\right)$ and reflectance $r$ etc. The LiDAR points are transformed info each camera image by a homogenous transformation and a projection. This process can be formulated as follows:
\begin{equation}
	idx = K_{i}T_{i}L_{xyz},
\end{equation}
where $K_{i}$ and $T_{i}$ are camera intrinsic matrix and homogenous transformation matrix corresponding to camera image $I_{i}$, respectively. $idx \in \mathbb{R}^{N\times 2}$ is the index (pixel coordinates) of LiDAR points $L$ on camera image $I_{i}$, where $N$ is the number of LiDAR points. The general transformation is given by $T_{cam\leftarrow lidar}$. For nuScenes dataset, the complete transformation to each camera is:
\begin{equation}
	T_{i} = T_{\left(cam\leftarrow ego_{i}\right)}
	T_{\left(ego_{i}\leftarrow g\right)}
	T_{\left(g\leftarrow ego_{s}\right)}
	T_{\left(ego_{s}\leftarrow lidar\right)},
\end{equation} 
with transforms: LiDAR frame to the ego-vehicle frame for the timestamp of the sweep $T_{\left(ego_{s}\leftarrow lidar\right)}$; ego frame to the global frame $T_{\left(g\leftarrow ego_{s}\right)}$; global frame to the ego-vehicle frame for the timestamp of the image $T_{\left(ego_{i}\leftarrow g\right)}$; and ego frame to the camera $T_{\left(cam\leftarrow ego_{i}\right)}$. After the LiDAR points are transformed to the camera coordinate, the corresponding camera matrix $K_{i}$ projects the points into the image $I_{i}$. Afterwards, the $w\times w$ (e.g., $3\times 3$) image context information of each projection point position is reshaped and concatenated to the corresponding LiDAR point. The concatenated points are fed into a UNet semantic segmentation sub-network (e.g., Cylinder3D \cite{zhou2020cylinder3d}) to obtain the coarse features $F_{coarse}$. 

\subsection{Offset Learning Stage}\label{sec:offset}
Although the methods of early-fusion and mid-fusion have achieved promising results in benchmark datasets, the performance of these methods is also limited because of the weak spatiotemporal synchronization problem between the LiDAR and cameras. To address the problem mentioned above, our proposed LIF-Seg predicts an offset between the projected LiDAR point and corresponding pixel. The predicted offset is used to compensate and update the position of projected point features, and then the aligned image semantic features are fused with the coarse features for better segmentation.

In this stage, as illustrated in Fig. \ref{fig:lifseg_offset_learning} and Algorithm \ref{alg:lifseg}, we first utilize an image semantic segmentation sub-network to obtain the high-level image semantic features $F_{image}$. Trade-off between efficiency and performance, we adopt DeepLabv3+ \cite{chen2018encoder} as our image segmentation sub-network to extract image features. Simultaneously, the coarse features $F_{coarse}$ are also projected into the image feature map and form a pseudo-image feature map $F_{points}$ with the same size as the image features. The feature map $F_{points}$ is further fused with the image semantic features $F_{image}$ to predict an offset between the projected LiDAR point and corresponding pixel. The predicted $\textit{Offset}$ can be used to compensate and update the position of the projected point in image features. Afterwards, according to the updated position, the image semantic features $F_{image}$ are back-projected into 3D space and generate the point-wise features  $F^{'}_{image}$. The point-wise image features  $F^{'}_{image}$ are used to fuse with the coarse features $F_{coarse}$ to improve the performance of LiDAR segmentation.

\begin{figure}[!t]
	\centering
	\includegraphics[width=0.98\linewidth]{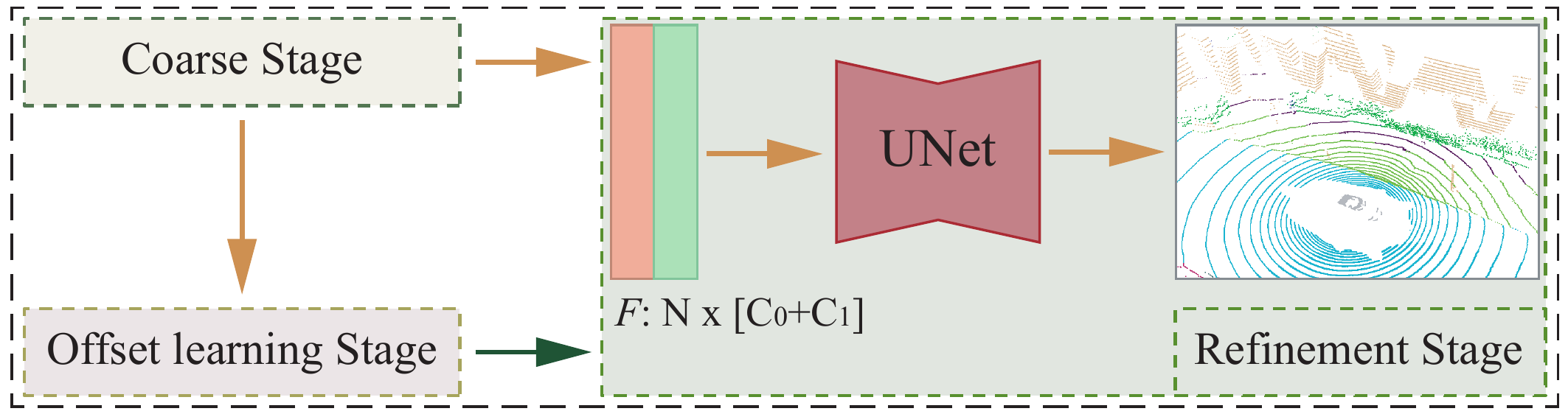}
	\caption{An overview of the connection between different stages and the illustration of the refinement stage.}
	\label{fig:lifseg_refinement}
\end{figure}

\subsection{Refinement Stage}
The refinement stage is illustrated in Fig. \ref{fig:lifseg_refinement}. After the coarse feature extraction stage and offset learning stage, we fuse the point-wise image features $F^{'}_{image}$ and the coarse features  $F_{coarse}$ by concatenation. Then, the concatenated features $F$ are fed into a UNet segmentation sub-network to obtain more accurate prediction results. For convenience, in the refinement stage, we use the same segmentation sub-network as in the coarse feature extraction stage. 

At the training time, we use a semantic segmentation loss $\mathcal{L}_{sem}$ to supervise the learning of LIF-Seg. The semantic segmentation loss $\mathcal{L}_{sem}$ consists of two items, including the classical cross-entropy loss and lovasz-softmax loss \cite{berman2018lovasz} to maximize the point accuracy and the intersection-over-union score, respectively. For the offset prediction in the subsection \ref{sec:offset}, taking the nuScenes \cite{caesar2020nuscenes} dataset as an example, there is no directly available supervision information for the offset learning because camera images corresponding to LiDAR point clouds do not provide pixel-level semantic or instance annotation. In this work, we utilize an auxiliary loss $\mathcal{L}_{aux}$ to supervise offset learning. Specifically, For points belonging to the foreground categories, we constrain their learned point-wise offset $O \in \mathbb{R}^{N\times 2}$ by a $L_{1}$ regression loss $\mathcal{L}_{reg}$:
\begin{equation}
	\mathcal{L}_{reg} = \frac{1}{\sum_{i}m_{i}}\sum_{i}\left\|o_{i} - \left(\hat{c}_{i} - p_{i}\right)\right\| \cdot m_{i},
\end{equation}
where $m = \{m_{1},\dots, m_{N}\}$ is a binary mask. $m_{i} = 1$ if point $i$ is in a 2D bounding box on image plane and $m_{i} = 0$ otherwise. $\hat{c}_{i}$ is the centroid of the 2D bounding box that point $i$ belongs to. Thus, the $\hat{c}_{i}$ can be formulated as follows:
\begin{equation}
	\hat{c}_{i} = \frac{1}{N^{B}_{g\left(i\right)}}\sum_{j\in B_{g\left(i\right)}}p_{j},
\end{equation}
where $g\left(i\right)$ maps point $i$ to the index of its corresponding 2D bounding box that contains point $i$. $N^{B}_{g\left(i\right)}$ is the number of points in 2D bounding box $B_{g\left(i\right)}$. To ensure these points move towards their corresponding centroid in the horizontal direction, we  utilize a direction loss $\mathcal{L}_{dir}$ to constrain the direction of predicted point-wise offset $O$. Following \cite{jiang2020pointgroup}, the $\mathcal{L}_{dir}$ is formulated as the average of minus cosine similarities:
\begin{equation}
	\mathcal{L}_{dir} = -\frac{1}{\sum_{i}m_{i}}\sum_{i}\frac{o_{i}}{\left\|o_{i}\right\|_{2}}\cdot\frac{\hat{c}_{i}-p_{i}}{\left\|\hat{c}_{i}-p_{i}\right\|_{2}}\cdot m_{i}.
\end{equation}
Thus, the auxiliary loss can be formulated as $\mathcal{L}_{aux} = \mathcal{L}_{reg} + \mathcal{L}_{dir}$. The training objective of our network is
\begin{equation}
	\mathcal{L} = \mathcal{L}_{sem} + \alpha\mathcal{L}_{aux},
\end{equation}
where $\alpha$ is the weight of auxiliary segmentation loss and set to $0.01$ in our experiments.

\section{Experiments}
In this section, we evaluate our approach on nuScenes \cite{caesar2020nuscenes} dataset to demonstrate the effectiveness of the proposed LIF-Seg. In the following, we first present a brief introduction to the dataset and evaluation metric in subsection \ref{sec:dataset_metric}. Then, the implementation details are provided in subsection \ref{sec:implementation}. Subsequently, we exhibit the detailed experiments about LiDAR-camera fusion and the comparisons with state-of-the-art methods on the nuScenes dataset in subsection \ref{sec:sota_results}. Finally, we conduct ablation studies to validate the effectiveness of offset learning in subsection \ref{sec:ablation_study}.

\subsection{Dataset and Evaluation Metric}\label{sec:dataset_metric}
The newly released nuScenes \cite{caesar2020nuscenes} dataset is a large-scale multi-modal dataset for LiDAR semantic segmentation, with more than 1000 scenes collected from different areas of Boston and Singapore. The scenes are split into 28,130 training frames and 6,019 validation frames. The annotated dataset provides up to 32 classes. After merging similar classes and removing rare classes, total 16 classes for the LiDAR semantic segmentation have remained. The dataset is collected by using a Velodyne HDL-32E sensor, cameras and radars with complete 360 coverage. In this work, we use the LiDAR point clouds and RGB images from all 6 cameras. Furthermore, this dataset has an imbalance challenge in different categories. In particular, classes like cars and pedestrians are most frequent, while bicycles and construction vehicles have relatively limited training data. Moreover, the nuScenes dataset is challenging as it is collected from different locations and diverse weather conditions. The point clouds of nuScenes are also less dense, because the sensor has fewer number of beams and lower horizontal angular resolution. 

\begin{table*}[!t]
	\caption{Experiment results of different early-fusion and mid-fusion on nuScenes validation set. RePr is our reproduced Cylinder3D.}
	\centering
	\resizebox{.98\linewidth}{!}
	{
		\begin{tabular}{c|c|cccccccccccccccc}
			\toprule
			Methods & $mIoU$ & \rotatebox{90}{barrier} & \rotatebox{90}{bicycle} & \rotatebox{90}{bus} & \rotatebox{90}{car} & \rotatebox{90}{construction} & \rotatebox{90}{motorcycle} & \rotatebox{90}{pedestrian} & \rotatebox{90}{traffic-cone} & \rotatebox{90}{trailer} & \rotatebox{90}{truck} & \rotatebox{90}{driveable} & \rotatebox{90}{other} & \rotatebox{90}{sidewalk} & \rotatebox{90}{terrain} & \rotatebox{90}{manmade} & \rotatebox{90}{vegetation} \\
			\midrule
			Cylinder3D (RePr) & 74.3  & 74.8  & 39.0  & 91.0  & 87.9  & 45.8  & 78.5  & 78.6  & 62.5  & 62.7  & 82.9  & 96.4  & 69.1  & 73.5  & 72.9  & 87.1  & 86.7 \\
			C+1$\times$1 & 74.8  & 75.9  & 37.7  & 89.6  & 88.3  & 50.9  & 79.4  & 78.9  & 68.2  & 60.1  & 82.7  & 96.4  & 69.8  & 72.9  & 71.2  & 87.9  & 87.1 \\
			C+3$\times$3 & 75.5  & 75.8  & 39.9  & 87.4  & 88.6  & 53.3  & 81.7  & 78.5  & 69.7  & 63.2  & 82.1  & 96.4  & 68.4  & 73.4  & 74.0  & 88.0  & \textbf{87.4} \\
			C+5$\times$5 & 75.0  & 75.1  & 36.4  & 90.6  & 88.4  & 54.8  & 77.5  & 78.7  & 68.0  & 59.1  & 82.4  & 96.3  & 70.4  & 73.6  & 74.2  & 88.0  & 87.2 \\
			C+Sem. & 75.7  & 75.5  & 45.4  & 91.3  & 87.5  & 50.4  & 83.4  & 81.3  & 67.9  & 61.4  & 81.0  & 96.3  & 68.7  & 73.4  & 74.3  & \textbf{88.6}  & 85.5 \\
			C+3$\times$3+Sem. & 76.4  & 74.7  & 46.2  & 91.9  & 87.2  & 50.5  & 79.2  & 80.6  & 70.7  & 66.4  & 82.7  & 96.3  & \textbf{72.8}  & 73.4  & \textbf{74.7}  & 88.2  & 86.9 \\ 
			\midrule
			C+Mid. & 74.8  & 72.9  & \textbf{49.9}  & 85.4  & 89.6  & 47.5  & 79.1  & 79.8  & 63.3  & 58.5  & 80.3  & 96.4  & 71.3  & 73.7  & 73.8  & \textbf{88.6}  & 87.2 \\
			\midrule
			C+3$\times$3+Mid. & 75.1 & 75.6 & 45.7 & 89.0 & \textbf{91.4} & 49.6 & 74.4 & 80.4 & 69.3 & 57.9 & 80.9 & 96.2 & 69.8 & 73.0 & 73.9 & 88.4 & 86.7 \\
			C+3$\times$3+Mid.+Ref. & \textbf{77.6} & \textbf{76.6} & \textbf{49.9} & \textbf{92.2} & 88.8 & \textbf{56.8} & \textbf{83.7} & \textbf{81.6} & \textbf{71.5} & \textbf{67.2} & \textbf{84.2} & \textbf{96.5} & 69.7 & \textbf{74.3} & 73.7 & 88.2 & 87.0 \\
			\bottomrule
		\end{tabular}
	}
	\label{tab:nuscenes_fusion}
\end{table*}

To evaluate the LiDAR semantic segmentation performance of our proposed approach, the mean intersection-over-union (mIoU) over all classes is taken as the evaluation metric. The mIoU can be formulated as 
\begin{equation}
	mIoU=\frac{1}{C}\sum^{C}_{i=1}IoU_{i}, 
\end{equation}
\begin{equation} 
	IoU_{i}=\frac{p_{ii}}{p_{ii} + \sum_{j \neq i}p_{ij} + \sum_{k\neq i}p_{ki}}, 
\end{equation}
where $C$ is the number of classes, and $p_{ij}$ denotes the number of points from class $i$ predicted as class $j$.

\subsection{Implementation Details}\label{sec:implementation}
\textit{Image Semantic Network Details.}\quad For the image semantic segmentation sub-network DeepLabV3+ \cite{chen2018encoder}, it takes a ResNet \cite{he2016deep} network as backbone to generate features at stride 16 and a FCN \cite{long2015fully} segmentation head to generate full-resolution semantic features $F_{image}\in\mathbb{R}^{n\times H\times W \times C_{1}}$, where $n=6$ is the number of cameras and $C_{1}=16$ is the dimension of features. However, there is no public segmentation pretrain model on nuScenes so we train the DeepLabV3+ \footnote{https://github.com/VainF/DeepLabV3Plus-Pytorch} by using the nuImages \footnote{https://www.nuscenes.org/images.} dataset. The nuImages consists of $100k$ images annotated with semantic segmentation labels. Note that all classes of nuImages are part of nuScenes. Moreover, the images of nuImages hardly exist in the image set corresponding to LiDAR point clouds of the nuScenes dataset.

\textit{LiDAR Network Details.} \quad For the LiDAR point clouds segmentation sub-network in coarse and refinement stages, we adopt Cylinder3D \cite{zhou2020cylinder3d} as the sub-network in these two stages. For the nuScenes dataset, cylindrical partition splits the LiDAR point clouds into 3D representation with the size $480\times 360 \times 32$, where three dimensions indicate the radius, angle and height, respectively. Besides, the feature dimension $C_{0}$ of coarse features $F_{coarse}$ is set to $C_{0}=C$, where $C$ is the number of categories. The window size $w$ of image contextual information is set to 3.


\subsection{Performance Results and Analyses}\label{sec:sota_results}
In this sub-section, we first conduct extensive experiments on the validation set of nuScenes \cite{caesar2020nuscenes} dataset to validate the effectiveness of different LiDAR-camera fusion strategies including early-fusion of between LiDAR and different context of the camera image, mid-fusion of between the LiDAR point features and image semantic features. Afterwards, we exhibit the comparisons with state-of-the-art methods on the nuScenes dataset. For all experiments, we adopt the retrained DeepLabV3+ \cite{chen2018encoder} to extract image features and the Cylinder3D \cite{zhou2020cylinder3d} to take as LiDAR segmentation baseline. For a fairer and clearer comparison, we retrain the baseline network Cylinder3D \footnote{https://github.com/xinge008/Cylinder3D} by using the code at GitHub published by the author, and if there are no extra notes, we use the same fusion strategy to fuse LiDAR and camera image in all models. 

\textit{Early-fusion and Mid-fusion.} \quad For the early-fusion, LiDAR points are projected into camera images by transformation matrixes and camera matrixes. According to the position of projected points, we can query the contextual information of the image with the window size $w\times w$ such as $1\times 1$, $3\times 3$ and $5\times 5$. The $w\times w$ contextual information is reshaped to a vector and concatenated to the corresponding LiDAR point. The concatenated points are fed into the baseline network Cylinder3D to obtain the segmentation results, and the models of different contextual information fusion are denoted as C+1$\times$1, C+3$\times$3 and C+5$\times$5, respectively. Besides, the channel-wise image semantic features obtained by DeepLabV3+ are also appended to each LiDAR point to enhance the point features (denoted as C+Sem.). Moreover, we also fuse the $3\times 3$ image contextual information and image semantic features in early-fusion (denoted as C+3$\times$3+Sem.). For the mid-fusion, the image semantic features are fused with LiDAR point features obtained by baseline network by concatenation (denoted as C+Mid.). The fused features are applied to two convolutional layers to generate the segmentation results. Besides, we also fuse the $3\times 3$ image contextual information in early-stage based on the mid-fusion method C+Mid. (denoted as C+3$\times$3+Mid.). Finally, Cylinder3D is also taken as a refinement sub-network to replace the two convolutional layers in C+3$\times$3+Mid. (denoted as C+3$\times$3+Mid.+Ref.).

\begin{table*}[!ht]
	\caption{Experiment results of our proposed method LIF-Seg and other LiDAR Segmentation methods on nuScenes validation set.} 
	\centering
	\resizebox{.95\linewidth}{!}
	{
		\begin{tabular}{c|c|cccccccccccccccc}
			\toprule
			Methods & $mIoU$ & \rotatebox{90}{barrier} & \rotatebox{90}{bicycle} & \rotatebox{90}{bus} & \rotatebox{90}{car} & \rotatebox{90}{construction} & \rotatebox{90}{motorcycle} & \rotatebox{90}{pedestrian} & \rotatebox{90}{traffic-cone} & \rotatebox{90}{trailer} & \rotatebox{90}{truck} & \rotatebox{90}{driveable} & \rotatebox{90}{other} & \rotatebox{90}{sidewalk} & \rotatebox{90}{terrain} & \rotatebox{90}{manmade} & \rotatebox{90}{vegetation} \\
			\midrule
			\#Points (k) & - & 1629 & 21 & 851 & 6130 & 194 & 81 & 417 & 112 & 370 & 2560 & 56048 & 1972 & 12631 & 13620 & 31667 & 21948 \\
			\midrule
			$(AF)^2$-S3Net \cite{cheng2021af} & 62.2 & 60.3 & 12.6 & 82.3 & 80.0 & 20.1 & 62.0 & 59.0 & 49.0 & 42.2 & 67.4 & 94.2 & 68.0 & 64.1 & 68.6 & 82.9 & 82.4 \\
			RangeNet++ \cite{milioto2019rangenet++} & 65.5 & 66.0 & 21.3 & 77.2 & 80.9 & 30.2 & 66.8 & 69.6 & 52.1 & 54.2 & 72.3 & 94.1 & 66.6 & 63.5 & 70.1 & 83.1 & 79.8 \\
			PolarNet \cite{zhang2020polarnet} & 71.0 & 74.7 & 28.2 & 85.3 & 90.9 & 35.1 & 77.5 & 71.3 & 58.8 & 57.4 & 76.1 & 96.5 & 71.1 & 74.7 & 74.0 & 87.3 & 85.7 \\
			Salsanext \cite{cortinhal2020salsanext} & 72.2 & 74.8 & 34.1 & 85.9 & 88.4 & 42.2 & 72.4 & 72.2 & 63.1 & 61.3 & 76.5 & 96.0 & 70.8 & 71.2 & 71.5 & 86.7 & 84.4 \\
			Cylinder3D \cite{zhou2020cylinder3d} & 76.1 & 76.4 & 40.3 & 91.2 & \textbf{93.8} & 51.3 & 78.0 & 78.9 & 64.9 & 62.1 & \textbf{84.4} & \textbf{96.8} & 71.6 & \textbf{76.4} & \textbf{75.4} & \textbf{90.5} & 87.4 \\ 
			LIF-Seg (Ours) & \textbf{78.2} & \textbf{76.5} & \textbf{51.4} & \textbf{91.5} & 89.2 & \textbf{58.4} & \textbf{86.6} & \textbf{82.7} & \textbf{72.9} & \textbf{65.5} & 84.1 & 96.7 & \textbf{73.2} & 74.4 & 73.1 & 87.5 & \textbf{87.6} \\
			\bottomrule
		\end{tabular}
	}
	\label{tab:nuscenes_lifseg}
\end{table*}

\begin{figure*}[!ht]
	\centering
	\includegraphics[width=0.95\linewidth]{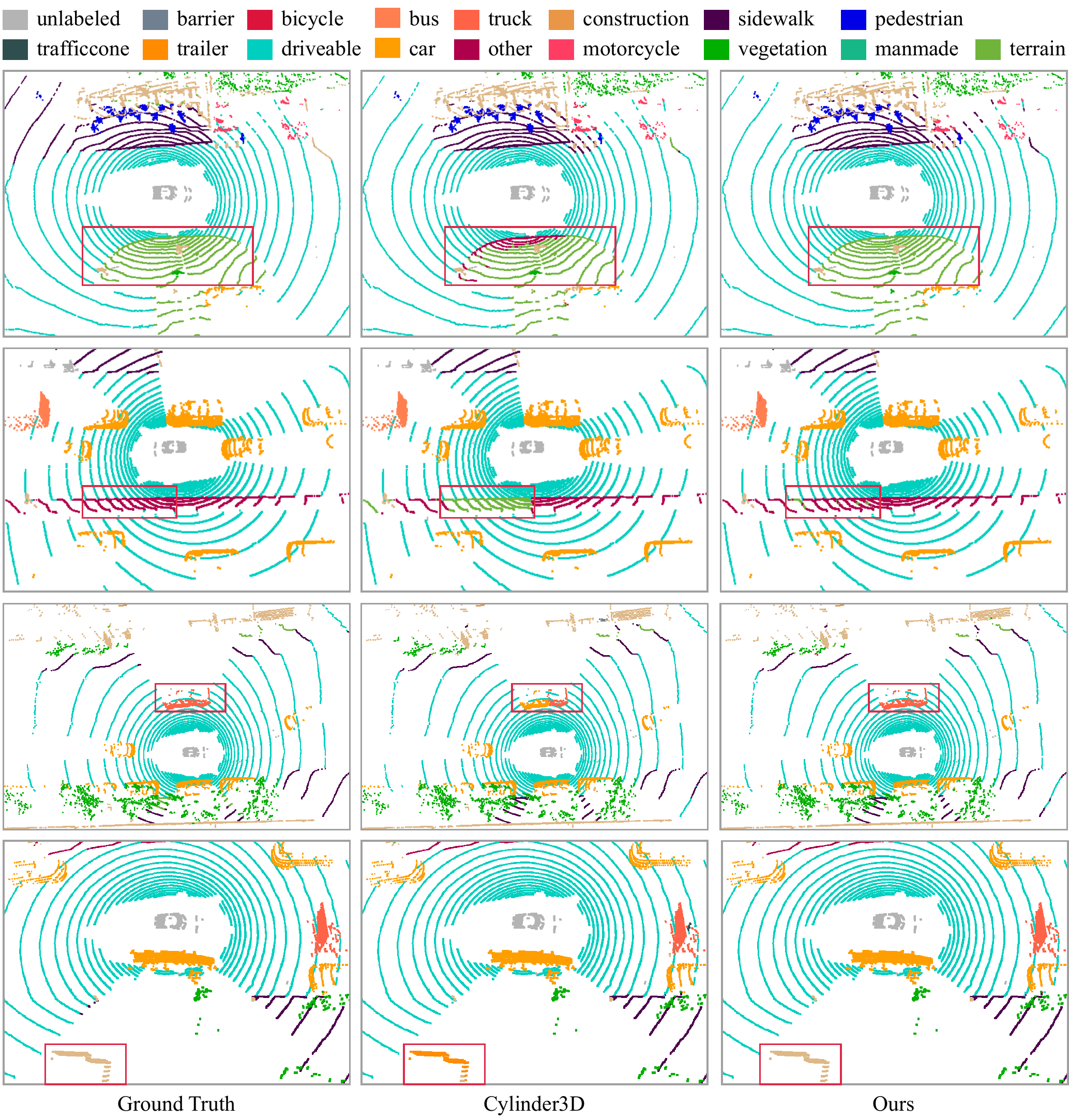}
	\caption{Comparison results of Cylinder3D and our method in LiDAR semantic segmentation tasks on nuScenes dataset validation set. Best viewed in color.}
	\label{fig:lifseg_results}
\end{figure*}

The LiDAR semantic segmentation results of different LiDAR-camera fusion strategies are depicted in Table \ref{tab:nuscenes_fusion}. Compared with the baseline method Cylinder3D and the C+1$\times$1, we can see that direct fusion of the LiDAR and image information can improve the performance of LiDAR semantic segmentation. Compared with the early-fusion methods C+1$\times$1, C+3$\times$3 and C+5$\times$5, the C+3$\times$3 achieves the best mIoU score because of the fusion image contextual information. The fusion method C+1$\times$1 lacks contextual information, limiting its ability to recognize fine-grained patterns. The context window size of fusion method C+5$\times$5 is too large, and too much redundant information limits the recognition of the central point semantic category. Similar to the 3D detector PointPainting \cite{vora2020pointpainting}, the early-fusion method C+Sem. can also improve the performance of LiDAR segmentation. Besides, the C+3$\times$3+Sem. indicates fusing the LiDAR points, image contextual information and semantic features can effectively improve the performance of semantic segmentation. The fusion methods C+Mid. and C+3$\times$3+Mid. are also slightly better than the baseline because of lacking the well-designed mid-fusion module. The experiment results of C+3$\times$3+Mid.+Ref. indicates well-designed mid-fusion module can effectively improve the performance of segmentation. These experimental results show that the image context information and image semantic features are helpful for LiDAR segmentation. In this work, LiDAR points and image contextual information are fused in the coarse stage, and the point features and aligned image semantic features are fused in the refinement stage.

\begin{table}[!t]
	\caption{Ablation results on nuScenes dataset validation set.}
	\centering
	\resizebox{.80\linewidth}{!}
	{
		\begin{tabular}{l|c}
			\toprule
			Method & $mIoU$ \\
			\midrule
			Remove the offset learning stage & 77.6 \\
			LIF-Seg (Ours) & \textbf{78.2} \\
			\bottomrule
		\end{tabular}
	}
	\label{tab:nuscenes_ablation}
\end{table}

\textit{Comparison with the SOTA Methods.} \quad Following \cite{zhou2020cylinder3d}, we conduct experiments on nuScenes \cite{caesar2020nuscenes} dataset to evaluate the effectiveness of our method. Table \ref{tab:nuscenes_lifseg} presents the LiDAR semantic segmentation results on nuScenes validation set. The RangeNet++ \cite{milioto2019rangenet++} and Salsanext \cite{cortinhal2020salsanext} perform the post-precessing. From Table \ref{tab:nuscenes_lifseg}, we can see that our proposed method achieves better performance than other methods and is dominant in many categories. Specifically, the proposed method outperforms Cylinder3D \cite{zhou2020cylinder3d} by 2.1 mIoU. Moreover, compared with the state-of-the-art projection-based methods (e.g., RangeNet++ and Salsanext), the LIF-Seg achieves about $6\%\sim12\%$ performance gain. Note that the points of nuScenes are very sparse (35k points/frame), especially for bicycles, motorcycles, traffic-cones, and pedestrians, etc. Therefore, the LiDAR segmentation task is more challenging. From Table \ref{tab:nuscenes_lifseg}, we can see that our method significantly outperforms other approaches in those sparse categories, because the LIF-Seg effectively fuses the LiDAR points, the camera image contextual information and image semantic features by a coarse-to-fine framework. Qualitative results of LiDAR segmentation are presented in Fig. \ref{fig:lifseg_results}.

\subsection{Ablation Studies}\label{sec:ablation_study} 
\begin{figure}[!t]
	\centering
	\includegraphics[width=0.98\columnwidth]{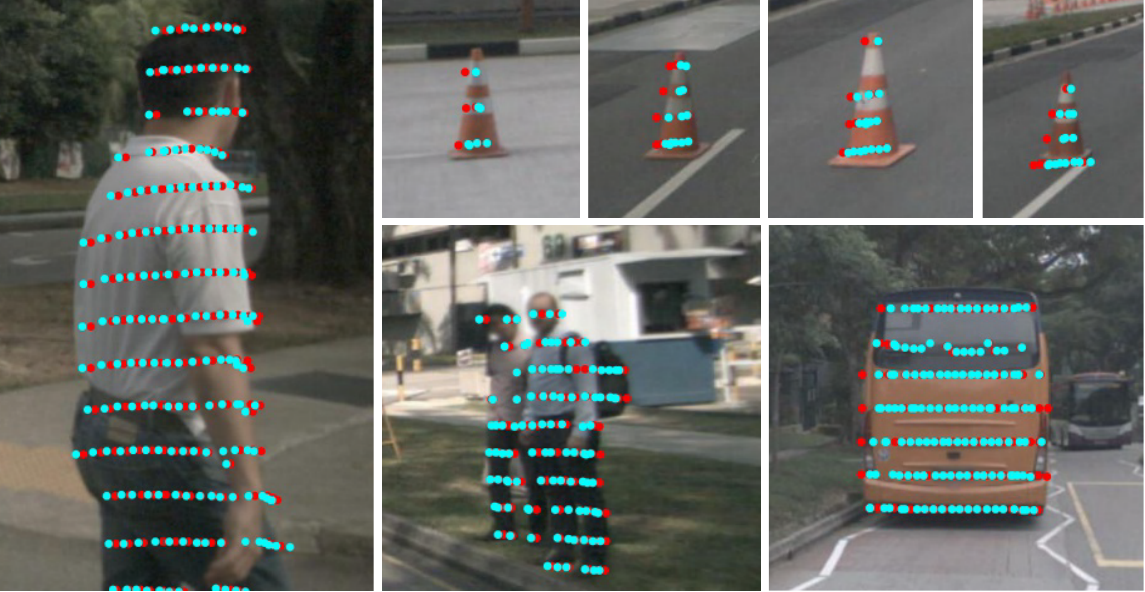}
	\caption{Example scenes from the nuScenes \cite{caesar2020nuscenes} dataset. The red point is the position of the original LiDAR point projected onto the camera image, and the cyan point is the updated position by using the predicted offset. }
	\label{fig:offset_vis}
\end{figure}

In this sub-section, we conduct ablation experiments on the validation set of nuScenes \cite{caesar2020nuscenes} dataset to validate the effectiveness of offset learning. For a fairer and clearer comparison, if there are no extra notes, we use the same configuration and sequential fusion strategy for all models. Detailed ablation experiments results are presented in Tabel \ref{tab:nuscenes_ablation}. We remove the offset learning stage from the full pipeline LIF-Seg, which causes the performance of LiDAR segmentation to drop from 78.2 to 77.6 mIoU. The offset prediction results are presented in Fig. \ref{fig:offset_vis}. From Fig. \ref{fig:offset_vis}, we can see that the projected points move towards their corresponding centroid in the horizontal direction, which makes these points fall on the instance object as much as possible. These results demonstrate the effectiveness of our method.

\section{Conclusion}
In this paper, we propose a coarse-to-fine framework LIF-Seg to improve the 3D semantic segmentation performance from two aspects including low-level image contextual information fusion in early-stage, and aligned high-level image semantic information fusion by tackling the weak spatiotemporal synchronization between the LiDAR and camera. The LIF-Seg consists of three main stages: coarse stage, offset learning stage and refinement stage. In the coarse stage, the LiDAR points and low-level image contextual information are fused and fed into a UNet sub-network to generate coarse features. The coarse features and image semantic features obtained by an image segmentation sub-network are fused to predict an offset between each projected LiDAR point and image pixel. The predicted offsets are used to align the coarse features and image semantic features. In the refinement stage, the coarse features and aligned image semantic features are fused and fed into a UNet sub-network to obtain more accurate semantic segmentation results. Extensive experimental results on nuScenes dataset demonstrate the effectiveness of our method. In the future, unsupervised learning methods can be added into our LIF-Seg to predict a transformation matrix between the LiDAR and camera to address the weak spatiotemporal synchronization problem completely and further improve the performance of LiDAR segmentation.

%

\ifCLASSOPTIONcaptionsoff
  \newpage
\fi


\bibliographystyle{IEEEtran}
\bibliography{bare_jrnl}

%

%
%
%




\end{document}